\documentclass[runningheads]{llncs}
\usepackage{graphicx}
\usepackage{amsmath,amssymb} 
\usepackage{color}
\usepackage[width=122mm,left=12mm,paperwidth=146mm,height=193mm,top=12mm,paperheight=217mm]{geometry}
\usepackage{floatrow, makecell}
\setcellgapes{0.1pt}
\usepackage{float}
\floatstyle{plaintop}
\restylefloat{table}

\begin{document}
\pagestyle{headings}
\mainmatter

\title{Adversarial Network Compression} 

\titlerunning{Adversarial Network Compression}

\authorrunning{Belagiannis et. al.}

\newcommand*\samethanks[1][\value{footnote}]{\footnotemark[#1]}

\author{Vasileios Belagiannis\thanks{Equal contribution.}\textsuperscript{1}, Azade Farshad\samethanks\textsuperscript{1,2}, Fabio Galasso\textsuperscript{1}}

\institute{Innovation OSRAM GmbH, Garching b. M{\"u}nchen, Germany
\and
Technische Universit{\"a}t M{\"u}nchen, Garching b. M{\"u}nchen, Germany
}

\maketitle
\begin{abstract}
Neural network compression has recently received much attention due to the computational requirements of modern deep models. In this work, our objective is to transfer knowledge from a deep and accurate model to a smaller one. Our contributions are threefold: (i) we propose an adversarial network compression approach to train the small student network to mimic the large teacher, without the need for labels during training; (ii) we introduce a regularization scheme to prevent a trivially-strong discriminator without reducing the network capacity and (iii) our approach generalizes on different teacher-student models.

In an extensive evaluation on five standard datasets, we show that our student has small accuracy drop, achieves better performance than other knowledge transfer approaches and it surpasses the performance of the same network trained with labels. In addition, we demonstrate state-of-the-art results compared to other compression strategies.
\end{abstract}

\section{Introduction}

Deep learning approaches dominate on most recognition tasks nowadays. Convolutional Neural Networks (ConvNets) rank highest on classification~\cite{xie2016aggregated}, object detection~\cite{FocalLoss2017}, image segmentation~\cite{chen2016deeplab} and pose estimation~\cite{newell2016stacked}, just to name a few examples. However, the superior performance comes at the cost of model complexity and large hardware requirements. Consequently, deep models often struggle to achieve real-time inference and cannot generally be deployed on resource-constrained devices, such as mobile phones.

In this work, our objective is to compress a large and complex deep network to smaller one. Network compression is a solution that only recently attracted more attention because of the deep neural networks. One can train a deep model with quantized or binarized parameters~\cite{soudry2014expectation,rastegari2016xnor,wu2016quantized}, factorize it, prune network connections~\cite{han2015learning,jaderberg2014speeding} or transfer knowledge to a small network~\cite{ba2014deep,buciluǎ2006model,hinton2015distilling}. In the latter case, the $student$ network is trained with the aid of the $teacher$.

We present a network compression algorithm whereby we complement the knowledge transfer, in the teacher-student paradigm, with adversarial training. In our method, a large and accurate $teacher$ ConvNet is trained in advance. Then, a small $student$ ConvNet is trained to mimic the $teacher$, i.e.\ to obtain the same output. Our novelty lies in drawing inspiration from Generative Adversarial Networks (GANs)~\cite{goodfellow2014generative} to align the teacher-student distributions. We propose a two-player game, where the discriminator distinguishes whether the input comes from the teacher or student, thus effectively pushing the two distributions close to each other. In addition, we come up with a regularization scheme to  help the student in competing with the discriminator. Our method does not require labels, only the discriminator's objectives and an L2 loss between the teacher and student output. We name our new algorithm \emph{adversarial network compression}.

An extensive evaluation on CIFAR-10~\cite{krizhevsky2009learning}, CIFAR-100~\cite{krizhevsky2009learning}, SVHN~\cite{netzersvhn}, Fashion-MNIST~\cite{xiao2017fashion} and ImageNet~\cite{deng2009imagenet} shows that our student network has small accuracy drop and achieves better performance than the related approaches on knowledge transfer. In addition, we constantly observe that our student achieves better accuracy than the same network trained with supervision (i.e.~labels). In our comparisons, we demonstrate superior performance next to other compression approaches. Finally, we employ three teacher and three student architectures to support our claim for generalization. 

We make the following contributions: (i) a knowledge transfer method based on adversarial learning to bridge the performance of a large model with a smaller one with limited capacity, without requiring labels during learning; (ii) a regularization scheme to prevent a trivially-strong discriminator and (iii) generalization on different teacher-student architectures.

\section{Related Work}

Neural network compression has been known since the early work of~\cite{hassibi1993second,strom1997phoneme}, but recently received much attention due to the combined growth of performance and computational requirements in modern deep models. Our work mostly relates to model compression~\cite{ba2014deep,buciluǎ2006model} and network distillation~\cite{hinton2015distilling}. We review the related approaches on neural network compression by defining five main categories and then discuss adversarial training.

\paragraph{I.~Quantization \& Binarization} The standard way to reduce the size and accelerate the inference is to use weights with discrete values~\cite{soudry2014expectation,wu2016quantized}. The Trained Ternary Quantization~\cite{zhu2016trained} reduces the precision of the network weights to ternary values. In incremental network quantization~\cite{zhou2017incremental}, the goal is to convert progressively a pre-trained full-precision ConvNet to a low-precision. Based on the same idea, Gong {\it et al}.~\cite{gong2014compressing} have clustered the weights using k-means and then performed quantization. The quantization can be efficiently reduced up to binary level as in XNOR-Net~\cite{rastegari2016xnor}, where the weight values are -1 and 1, and in BinaryConnect~\cite{courbariaux2015binary}, which binarizes the weights during the forward and backward passes but not during the parameters' update. Similar to binary approaches, ternary weights (-1,0,1) have been employed  as well~\cite{li2016ternary}.

\paragraph{II.~Pruning} Reducing the model size (memory and storage) is also the goal of pruning by removing network connections~\cite{carreira2018learning,tung2018clip,yu2017nisp}. At the same time, it prevents over-fitting. In~\cite{han2015learning}, the unimportant connections of the network are pruned and the remaining network is fine-tuned. Han {\it et al.}~\cite{han2015deep} have combined the idea of quantization with pruning to further reduce the storage requirements and network computations. In HashedNets~\cite{chen2015compressing}, the network connections have been randomly grouped into hash buckets where all connections of the same bucket share the weight. However, the sparse connections do not necessarily accelerate the inference when employing ConvNets. For this reason, Li {\it et al.}~\cite{li2016pruning} have pruned complete filters instead of individual connections. Consequently, the pruned network still operates with dense matrix multiplications and it does not require sparse convolution libraries. Parameter sharing has also contributed to reduce the network parameters in neural networks with repetitive patterns~\cite{belagiannis2017recurrent,schmidhuber1992learning}.

\paragraph{III.~Decomposition / Factorization} In this case, the main idea is to construct low rank basis of filters. For instance, Jaderberg {\it et al.}~\cite{jaderberg2014speeding} have proposed an agnostic approach to have rank-1 filters in the spatial domain. Related approaches have also explored the same principle of finding a low-rank approximation for the convolutional layers~\cite{rigamonti2013learning,denton2014exploiting,lebedev2014speeding,yang2015deep}. More recently, it has been proposed to use depthwise separable convolutions, as well as, pointwise convolutions to reduce the parameters of the network. For example, MobileNets are based on depthwise separable convolutions, followed by pointwise convolutions~\cite{howard2017mobilenets}. In a similar way, ShuffleNet is based on depthwise convolutions and pointwise group convolutions, but it shuffles feature channels for increased robustness~\cite{zhang2017shufflenet}.

\paragraph{IV.~Efficient Network Design} The most widely employed deep models, AlexNet \cite{krizhevsky2012imagenet} and VGG16~\cite{simonyan2014very}, demand large computational resources. This has motivated more efficient architectures such as the Residual Networks (ResNets)~\cite{he2016deep} and their variants~\cite{huang2016deep,zagoruyko2016wide}, which reduced the parameters, but maintained (or improved) the performance. SqueezeNet~\cite{iandola2016squeezenet} trims the parameters further by replacing 3x3 filters with  1x1 filters and decreasing the number of channels for 3x3 filters. Other recent architectures such as Inception~\cite{szegedy2016rethinking}, Xception~\cite{chollet2016xception}, CondenseNet~\cite{huang2017condensenet} and ResNeXt~\cite{xie2016aggregated} have also been efficiently designed to allow deeper and wider networks without introducing more parameters than AlexNet and VGG16. Among the recent architectures, we pick ResNet as the standard model to build the $teacher$ and $student$. The reason is the model simplicity where our approach applies to ResNet variations as well as other architectures.

Network compression categories \textbf{I-IV} address the problem by reducing the network parameters, changing the network structure or designing computationally efficient components. By contrast, we focus on knowledge transfer from a complex to a simpler network without interventions on the architecture. Our knowledge transfer approach is closer to network distillation, but it has important differences that we discuss bellow.

\paragraph{V.~Distillation} Knowledge transfer has been successfully accomplished in the past~\cite{ba2014deep,li2014learning}, but it has been popularized by Hinton {\it et al.}~\cite{hinton2015distilling}. The goal is to transfer knowledge from the $teacher$ to the $student$ by using the output before the softmax function (logits) or after it (soft targets). This task is known as network compression~\cite{ba2014deep,buciluǎ2006model} and  distillation~\cite{hinton2015distilling,polino2018model}. In our work, we explore the problem for recent ConvNet architectures for the $teacher$ and $student$ roles. We demonstrate that network compression performs well with deep models, similar to the findings of Urban {\it et al.}~\cite{urban2016deep}. Differently from the earlier works, we introduce adversarial learning into compression for the first time, as a tool for transferring knowledge from the teacher to the student by their cooperative exploration. Also differently from the original idea of Hinton {\it et al.}~\cite{hinton2015distilling} and the recent one from Xu {\it et al.}~~\cite{xutraining}, we do not require labels for the $student$ training during compression. In the experiments, our results are constantly better than network distillation.

\paragraph{Adversarial Learning} Our work is related to the Generative Adversarial Networks (GAN)~\cite{goodfellow2014generative} where a network learns to generate images with adversarial learning, i.e.\ learning to generate images which cannot be distinguished by a discriminator network.
We take inspiration from GANs and introduce adversarial learning in model compression by challenging the student's output to become identical to the teacher. Closer to our objective is the work from Isola {\it et al.}~\cite{isola2016image} to map an image to another modality with a conditional Generative Adversarial Network (cGAN)~\cite{mirza2014conditional}. Although, we do not have a generator in our model, we aim to map the $student$ to $teacher$ output given the same input image. However, compared to the $teacher$, the $student$ is a model with limited capacity. This motivates a number of novel contributions, needed for the successful adversarial training. 

\section{Method}\label{sec:method}
We propose the adversarial network compression, a new approach to transfer knowledge between two networks. In this section, we define the problem and discuss our approach.

\subsection{Knowledge Transfer}
We define a deep and accurate network as $teacher$ $f_{t}(x;w_{t})$ and a small network as $student$ $f_{s}(x;w_{s})$. The $teacher$ has very large capacity and is trained on labeled data. The \textit{student} is a shallower network with significant less parameters. Both networks perform the same task, given an input image $x$. Our objective is to train the $student$ to mimic the $teacher$ by predicting the same output. To achieve it, we introduce the discriminator $D$, another network that learns to detect the $teacher$ / $student$ output based on adversarial training. We train the $student$ together with $D$ by using the knowledge of the $teacher$ for supervision.

\begin{figure*}[h!]
\centering
\begin{tabular}{c}
\includegraphics[scale=0.4, angle=0]{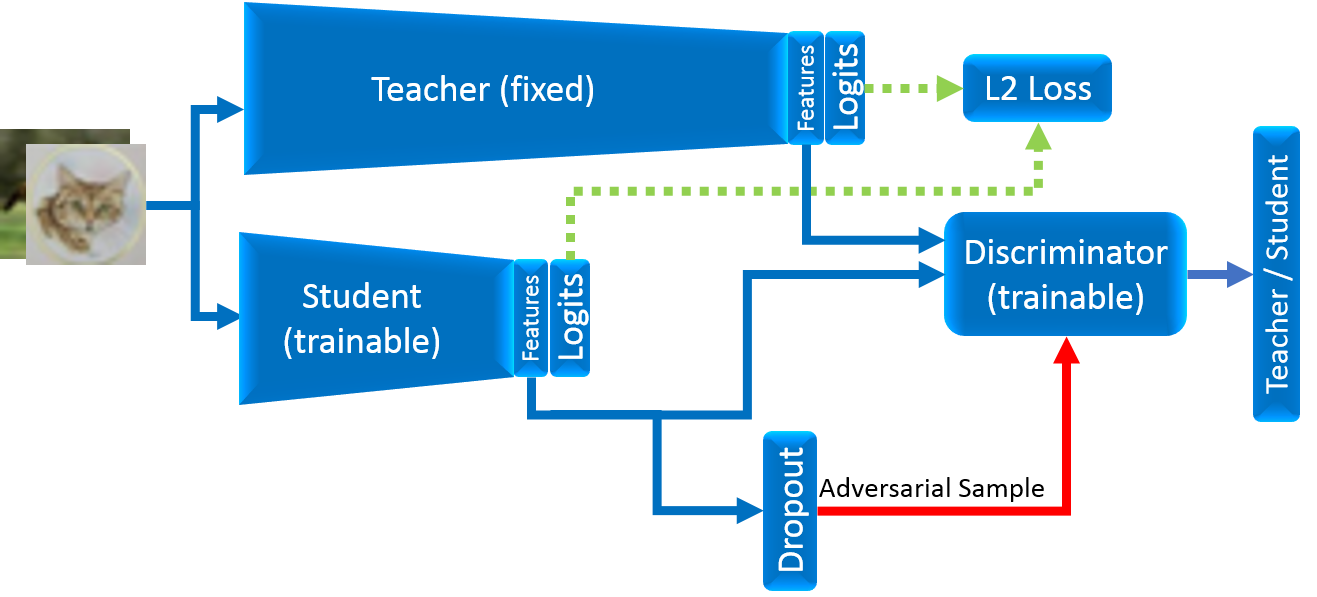}
\end{tabular}
\caption{\small {\bf Adversarial Network Compression}: Our method consists of the $teacher$, $student$ and discriminator networks. The teacher is trained in advance and used for supervision during adversarial learning, while the $student$ and discriminator are both trainable. The discriminator takes as input the features from the $teacher$ and $student$, as well as, the adversarial sample (i.e.~student labeled as teacher). For the adversarial sample, we empirically found that dropout is beneficial. In addition, there is a L2 loss to force the  $student$ to mimic the output of the $teacher$.}
\label{fig:Method}
\end{figure*}

In this work, we address the problem of classification. For transferring knowledge, we consider the unscaled log-probability values (i.e.~logits) before the softmax activation function, as well as, features from earlier layers. Bellow, we simplify the notation to $f_{t}(x)$ for the teacher and $f_{s}(x)$ for the student network output (logits). In addition, the feature representation of the $teacher$ at $k-th$ layer is defined as $f^{k}_{t}(x)$ and for the $student$ at $l-th$ layer is denoted as $f^{l}_{s}(x)$. In practice, $f^{k}_{t}(x)$ and $f^{l}_{s}(x)$ are the last layers before the logits. An overview of our method is illustrated in Fig.~\ref{fig:Method}.

\subsection{Generative Adversarial Networks}\label{ganTheory}
We shortly discuss the Generative Adversarial Networks (GANs)~\cite{goodfellow2014generative} to illustrate the connection with our approach. In GANs, the main idea is to simultaneously train two networks (two-player game) that compete with each other in order to improve their objectives. The first network, the generator $G$, takes random noise (i.e.~latent variables) input $z$ to generate images. In addition to the noise, the input can be conditioned on images or labels~\cite{mirza2014conditional}. In both cases, the goal is to learn generating images that look real by aligning the real data and model distributions. The second network, the discriminator $D$, takes as input an image from the data distribution and the generator's output; and the objective is to classify whether the image is real or fake. Overall, $G$ tries to fool $D$, while $D$ tries to detect input from $G$. Isola {\it et al.}~\cite{isola2016image} have shown that a conditional Generative Adversarial Network (cGAN) successfully transforms an input image to another modality $y$ using the adversarial learning. The objective of cGAN can be written as:
\begin{align}
    \mathcal{L}_{cGAN}(G,D) = &\mathbb{E}_{x,y \sim p_{data}(x,y)}[\log (D(x,y))] + \nonumber \\
                 &\mathbb{E}_{x \sim p_{data}(x), z \sim p_{z}(z)}[\log (1-D(x,G(x,z))],\label{cGANeq}
\end{align}
where $p_{data}(x,y)$ corresponds to the real data distribution over the input image $x$ and label $y$; and $p_{z}(z)$ to the prior distribution over the input noise $z$. During training, the objective is maximized w.r.t.~$D$ and minimized w.r.t.~$G$. In both cases, the loss is cross entropy for the binary $D$ output.

\subsection{Adversarial Compression}\label{AdvLearn}
In the context of network compression, we propose to adapt the two-player game based on the $teacher$ and $student$. The goal now is to adversarially train $D$ to classify whether input samples come from the $teacher$ or $student$ network. Both networks share the same input, namely an image $x$, but their predictions differ. We choose $f^{k}_{t}(x)$ and $f^{l}_{s}(x)$ feature representations from both networks as input to $D$. During training, $D$ is firstly updated w.r.t the labels from both input samples (blue lines, Fig.~\ref{fig:Method}). Next, the $student$ network is updated by inverting the labels for the $student$ samples (calling them $teacher$). The reason for changing the labels is to back-propagate gradients that guide the $student$ to produce output as the $teacher$ for the same input image (red line, Fig.~\ref{fig:Method}). The $teacher$ network has been trained in advance with labels and it is not updated during training. Eventually, fooling the discriminator translates in predicting the same output for $teacher$ and $student$ networks. This is the same objective as in network distillation. After reformulating Eq.~\eqref{cGANeq}, we define our objective as:
\begin{align}
\mathcal{L}_{Adv}(f_{s},D) = &\mathbb{E}_{f^{k}_{t}(x)\sim p_{teacher}(x)}[\log (D(f^{k}_{t}(x)))] + \nonumber \\
                &\mathbb{E}_{f^{l}_{s}(x) \sim p_{student}(x),  z \sim p_{z}(z)}[\log (1-D(f^{l}_{s}(x)))].\label{cAdveq}
\end{align}
where $p_{teacher}(x)$ and $p_{student}(x)$ correspond to the $teacher$ and $student$ feature distribution. We provide the noise input in the form of dropout applied on the $student$, similar to~\cite{isola2016image}. However, the dropout is active only during the student update (red arrow in Fig.~\ref{fig:Method}). We experimentally found that using the dropout only for the $student$ update gives more stable results for our problem. We omit $z$ in $f^{l}_{s}(x)$ to keep the notation simple.

The reason for using the features $f^{k}_{t}(x)$ and $f^{l}_{s}(x)$ as input to $D$, instead of the logits $f_{t}(x)$ and $f_{s}(x)$, is their dimensionality. The features usually have higher dimensions, which makes the judgment of the discriminator more challenging. We evaluate this statement later in the experimental section. Training $D$ with input from intermediate output ($f^{k}_{t}(x)$ and $f^{l}_{s}(x)$) from $teacher$ and $student$ works fine for updating the parameters of $D$ and partially for the $student$. In $student$, there is a number of parameters until the final output $f_{s}(x)$ (logits) which also has to be updated. To address this problem, we seek to minimize the difference between the two networks output, namely $f_{t}(x)$ and $f_{s}(x)$. This is the data objective in our formulation that is given by:
\begin{align}
\mathcal{L}_{Data}(f_{s}) = &\mathbb{E}_{f_{s}(x)\sim p_{student}(x)}\left[\parallel f_{t}(x) - f_{s}(x) \parallel^{2}_{2}\right]\label{cDateq}.
\end{align}
The data term contributes to the update of all $student$ parameters. We found it very important for the $student$ network convergence (green dashed lines, Fig.~\ref{fig:Method}) since our final goal is to match the output between the $teacher$ and $student$. The final objective with both terms is expressed as:
\begin{align}
\arg \min_{f_{s}}\max_{D} \mathcal{L}_{Adv}(f_{s},D) + \lambda \mathcal{L}_{Data}(f_{s}).\label{full_objective}
\end{align}
where $\lambda$ is a tuning constant between the two terms.

The data term of our model is the same with the compression objective of Ba and Caruana~\cite{ba2014deep} and in accordance with the work of Isola {\it et al}~\cite{isola2016image}. In addition, we aim for the exact output between $teacher$ - $student$ and thus using only the adversarial objective does not force the $student$ to be as close as possible to the $teacher$. Note also that the role of $G$ is implicitly assigned to the $student$, but it is not explicitly required in our approach. For the adversarial part, we share the label inversion idea for the adversarial samples from ADDA~\cite{tzeng2017adversarial} and reversal gradient~\cite{ganin2015unsupervised}. Below, we discuss the network architectures for exploring the idea of adversarial learning in network compression.

\subsection{Network Architectures}
Our model is composed of three networks: the $teacher$, $student$ and discriminator $D$. Here, we present all three architectures.

\textbf{Teacher}
We choose the latest version of ResNet~\cite{he2016identity} for this role, since it is currently the standard architecture for recognition tasks. The network has adaptive capacity based on the number of bottlenecks and number of feature per bottleneck. We select ResNet-164 for our experiments. To examine the generalization of our approach on small-scale experiments, the Network in Network (NiN)~\cite{lin2013network} is also selected as $teacher$. The $teacher$ is trained in advance with labeled data using cross-entropy.

\textbf{Student} We found it meaningful to choose ResNet architecture for the $student$ too. Although, the student is based on the same architecture, it has limited capacity. We perform our experiments with ResNet-18 and ResNet-20. For small-scale experiments, we employ LeNet-4~\cite{lecun1990handwritten} for the $student$ role. It is a shallow network and we experimentally found that it can be easily paired with NiN. The $student$ network parameters are not trained on the labeled data. Furthermore, the labels are not used in the adversarial compression.

\textbf{Discriminator} This discriminator $D$ plays the most important role among the others. It can be interpreted as a loss function with parameters. The discriminator has to strike a balance between simplicity and network capacity to avoid being trivially fooled. We choose empirically a relative architecture. Our network is composed of three fully-connected (FC) layers (128 - 256 -128) where the network input comes from the $teacher$ $f^{k}_{t}(x)$ and $student$ $f^{l}_{s}(x)$. The intermediate activations are non-linearities (ReLUs). The output is a binary prediction, given by a sigmoid function. The network is trained with cross entropy where the objective is to predict between $teacher$ or $student$. This architecture has been chosen among others, which we present in the experimental part (Sec.~\ref{sec:disArch}). Similar architectures are also maintained by \cite{tzeng2017adversarial} and~\cite{zhang2018unreasonable} for adversarial learning.

\subsection{Discriminator Regularization}\label{WeakD}
The input to the discriminator has significantly lower dimensions in our problem compared to GANs for image generation~\cite{isola2016image,salimans2016improved}. As a result, it is simpler for the discriminator to understand the source of input. In particular, it can easily distinguish $teacher$ from $student$ samples from the early training stages (as also maintained in~\cite{goodfellow2014generative}). To address this limitation, we explore different ways of regularizing the discriminator. Our goal is to prevent the discriminator from dominating the training, while retaining the network capacity. We consider therefore three types of regularization, which we examine in our experiments.

\textbf{L2 regularization} This is the standard way of regularizing a neural network~\cite{ng2004feature}. At first, we try the L2 regularization to force the weights of the discriminator not to grow. The term is given by:
\begin{align}
\mathcal{L}_{regul}(D) = - \mu \sum_{i=1}^{n} \left \| w_{D,i} \right \|^{2}_{2}\label{regL2}
\end{align}
where $n$ is the number of network parameters and $\mu$ controls the contribution of the regularizer to the optimization. The parameters of $D$ correspond to $w_{D,i}$.

\textbf{L1 regularization} Additionally, we try L1 regularization which supports sparse weights. This is formalized as:
\begin{align}
\mathcal{L}_{regul}(D) = - \mu \sum_{i=1}^{n} \left | w_{D,i} \right |\label{regL1}.
\end{align}
In both Eq.~\eqref{regL2} and \eqref{regL1} there is negative sign, because the term is updated during the maximization step of Eq.~\eqref{full_objective}.

\textbf{Adversarial samples for \emph{D}} In the above cases, there is no guarantee that the discriminator will become weaker w.r.t $student$. We propose to achieve it by updating $D$ with adversarial samples. According to the objective in Eq.~\eqref{full_objective}, the discriminator is updated only with correct labels. Here, we additionally update $D$ with $student$ samples that are labeled as $teacher$. This means that we use the same adversarial samples to update both $student$ and $D$. The new regularizer is defined as:
\begin{align}
\mathcal{L}_{regul}(D) = &\mathbb{E}_{f^{l}_{s}(x)\sim p_{student}(x)}[\log D(f^{l}_{s}(x))].\label{regul}
\end{align}
The motivation behind the regularizer is to prolong the game between the $student$ and discriminator. Eventually, the discriminator manages to distinguish $teacher$ and $student$ samples, as we have observed. However, the longer it takes the discriminator to win the game, the more valuable gradient updates the $student$ receives. The same principle has been also explored for text synthesis~\cite{reed2016generative}. Applying the same idea on the teacher samples does not hold, since it is fixed and thus a reference in training. Our objective now becomes:
\begin{align}
\arg \min_{f_{s}}\max_{D} \mathcal{L}_{Adv}(f_{s},D) + \lambda \mathcal{L}_{Data}(f_{s}) + \mathcal{L}_{regul}(D)\label{full_objective2}
\end{align}
where the regularization $\mathcal{L}_{regul}(D)$ corresponds to one of the above approaches. In the experimental section, we show that our method requires the regularization in order to achieve good results. In addition, we observed that the introduced regularization had the most significant influence in fooling $D$, since it is conditioned on the $student$.

\subsection{Learning \& Optimization}
The network compression occurs in two phases. First, the $teacher$ is trained from scratch on labeled data. Second, the $student$ is trained together with $D$. The $student$ is randomly initialized, as well as, $D$. All models are trained using Stochastic Gradient Descent (SGD) with momentum. The learning rate is $0.001$ for the first $80k$ training iterations and then it is decreased by one magnitude. The weight decay is set to $0.0002$. The min-batch size is to $128$ samples. Furthermore, the dropout is set to $0.5$ for the adversarial sample input to $D$. To further regularize the data, data augmentation (random crop and flip) is also included in training. In all experiments, the mean of the training set images is subtracted and they are divided by the standard deviation. Lastly, different weighting factors $\lambda$ have been examined, but we concluded that equal weighting is a good compromise for all evaluations. In the L1/L2 regularization, the value of $\mu$ is set to 0.99. The same protocol is followed for all datasets, unless it is differently stated. 

\section{Evaluation}\label{sec:results}

In this section, we evaluate our approach on five standard classification datasets: CIFAR-10~\cite{krizhevsky2009learning}, CIFAR-100~\cite{krizhevsky2009learning}, SVHN~\cite{netzersvhn}, Fashion-MNIST~\cite{xiao2017fashion} and ImageNet 2012~\cite{deng2009imagenet}. In total, we examine three $teacher$ and three $student$ architectures.

Our ultimate goal is to train the shallower and faster $student$ network to perform, at the level of accuracy, as close as possible to the deeper and complex $teacher$. Secondly, we aim to outperform the $student$ trained with supervision by transferring knowledge from the teacher. We report therefore for each experiment the error, numbers of parameters and floating point operations (FLOPs). The last two metrics are reported in M-Million or B-Billion scale.

Once we choose the discriminator $D$ and regularization in Sec.~\ref{sec:disArch}, we perform a set of baseline evaluations and comparisons with related approaches for all datasets in Sec.~\ref{sec:baselinesds} and Sec.~\ref{sec:imagenet}.

\textbf{Implementation details} Our implementation is based on TensorFlow~\cite{abadi2016tensorflow}. We also rely on our own implementation for the approach of Ba and Caruana~\cite{ba2014deep} and  Hinton {\it et al.}~\cite{hinton2015distilling}. The results of the other approaches are obtained from the respective publications. Regarding the network architectures, we rely on the official TF code for all ResNet variants, while we implement by ourselves the Network in Network (NiN) and LeNet-4 models.

\begin{table*}[h]
\centering
\caption{{\small {\bf Discriminator Evaluation}} We choose the discriminator which enables the best student performance on CIFAR-100, when integrated in the proposed adversarial compression framework. Fully connected (fc) and convolutional (conv) layers are examined. We report the $student$ classification error. The best performing model (128fc - 256fc - 128fc) is used in all other evaluations.}
\begin{tabular}{l|c|l|c}
\hline
Architecture & Error[\%]&Architecture & Error[\%]\\\hline
\textbf{128fc - 256fc - 128fc} & \textbf{32.45} & 500fc - 500fc & 33.28 \\
64fc - 128fc - 256fc & 32.78 & 256fc - 256fc - 64fc & 33.46 \\
256fc - 256fc & 32.82 & 64fc - 64fc & 33.51\\
256fc - 128fc - 64fc & 33.05 & 128conv - 256conv & 33.68 \\
64fc - 128fc - 128fc - 64fc & 33.09 & 128fc - 128fc - 128fc & 33.72 \\
\end{tabular}
\label{tab:Discri}
\end{table*}

\subsection{Discriminator Model}\label{sec:disArch}
We discuss the choice of the discriminator $D$ architecture and the impact of the regularization on $D$.

\textbf{Architecture} We examine which $D$ architecture should be considered for adversarial compression. To this end, we consider CIFAR-100 dataset as the most representative among the small-scale datasets and train our student with adversarial compression. Since the role of the discriminator would be to ensure the best student training, we explore several architectures and select the one that is providing the minimum classification error of the trained student. The discriminator models, except one, are fully connected (fc) with (ReLU) activation, other than the last layer. We explore two to four fc-layer models with different capacity. We also made experiments with a convolutional (conv) discriminator which has lower performance than fc discriminators. The results for the discriminator trials are in Table~\ref{tab:Discri}. The best architecture is given by 3 fc-layers of depth 128-256-128. Notice that  our best architecture is similar to the $D$ models for adversarial domain adaption~\cite{tzeng2017adversarial} and perceptual similarity~\cite{zhang2018unreasonable}.

\textbf{Regularization}\label{sec:regul}
Here we experiment on the three regularization techniques, described in Sec.~\ref{WeakD}, on four datasets. We rely on our best performing model, i.e.~the one with features provided as input to $D$ and dropout on the $student$. The results are summarized in Table~\ref{tab:Regul}. The lack of regularization leads to poorer performance since it is more difficult to fool the discriminator based on our observations. In particular, the performance without regularization is worse than training the $student$ architecture on supervised learning as we show in Sec.~\ref{sec:baselinesds}. Adding the L1 or L2 regularizer indicates an important error drop (L1 and L2 column in Table~\ref{tab:Regul}). However, our proposed regularization introduces the most difficulties in the discriminator that leads to better performance. We use the adversarial samples for $D$ regularization for the rest evaluations.

\subsection{Component Evaluation}\label{sec:baselinesds}
Initially, the $teacher$ is trained with labels (i.e.~supervised $teacher$ in Table~\ref{tab:CIFAR10}-\ref{tab:FMNIST}). Next, the adversarial compression is performed under different configurations. The input to $D$ is either the logits or features. In both cases, we also examine the effect of the dropout on the student. In all experiments, we train the student only based on teacher supervision and without labels. The results for every experiment are reported in Table~\ref{tab:CIFAR10}-\ref{tab:FMNIST}. In all baselines, there is the $L2$ loss on the logits from the $teacher$ and $student$ (i.e.~$f_{t}(x)$ and $f_{s}(x)$). We also provide the results of the same network as the $student$ trained with labels (i.e.~supervised $student$ in Table~\ref{tab:CIFAR10}-\ref{tab:FMNIST}).

\begin{table*}[h]
\centering
\caption{{\small {\bf Regularization Evaluation}} We evaluate three different ways of regularizing the discriminator $D$. We also show the performance without regularization. The error is in percentage. Our adversarial sample in $D$ regularization is presented in the last column. All experiments have been accomplished with our complete model, namely features input to $D$ and dropout on the $student$.}
\begin{tabular}{l|c|c|c|c|c|c}
Dataset   & Teacher    & Student& W/o Regul.& L1    & L2    & Ours  \\ \hline
CIFAR-10  & ResNet-164 & ResNet-20 &  10.07& 8.19 & 8.16 & 8.08 \\
CIFAR-100 & ResNet-164 & ResNet-20 & 34.10 & 33.36 & 33.02 & 32.45 \\
SVHN      & ResNet-164 & ResNet-20 & 3.73 & 3.67 & 3.68 & 3.66 \\
F-MNIST   & NiN        & LeNet-4   & 9.62 & 8.91 & 8.75 & 8.64
\end{tabular}
\label{tab:Regul}
\end{table*}

On CIFAR-10, CIFAR-100 and SVHN experiments, the input to $D$ from ResNet-164 and ResNet-20 is the features of the \textit{average pool} layer, which are used for $teacher$ $f^{k}_{t}(x)$ and $student$ $f^{l}_{s}(x)$. On Fashion-MNIST, it is the output of the last fully connected layer before the logits both for NiN (teacher) and LeNet-4 (student). The model training runs for $260$ epochs in CIFAR-10, CIFAR-100 and SVHN, while for $120$ epochs in Fashion-MNIST. Below, the results are individually discussed for each dataset.

\begin{table*}[h!]
  \floatsetup{floatrowsep=qquad, captionskip=4pt}
  \begin{floatrow}[2]
    \makegapedcells
    \ttabbox%
	{\begin{tabular}{l|c}
	\hline
	Model & Error[\%]\\\hline
	Supervised $teacher$& 6.57\\
    ResNet-164&\\\hline
	Supervised $student$& 8.58\\
    ResNet-20&\\\hline
	Our $student$ ($D$ with logits)& 8.31\\
    \quad\quad\quad+ dropout on $student$& 8.10\\
    Our $student$ ($D$ with features)& 8.10\\
    \quad\quad\quad+ dropout on $student$& 8.08\\
    \end{tabular}}
    {\caption{{\small {\bf CIFAR-10 Evaluation} We evaluate the components of our approach. ResNet-164 Parameters: \textbf{2.6M}, FLOPs: \textbf{97.49B}. ResNet-20 Parameters: \textbf{0.27M}, FLOPs: \textbf{10.52B}. Our $student$, ResNet-20, has around 10x less parameters.}}\label{tab:CIFAR10}}
    \hfill%
    \ttabbox%
    {\begin{tabular}{l|c}
    \hline
    Model & Error[\%]\\\hline
    Supervised $teacher$& 27.76\\
    ResNet-164& \\\hline
    Supervised $student$& 33.36\\
    ResNet-20& \\\hline
    Our $student$ ($D$ with logits)& 33.96\\
    \quad\quad\quad+ dropout on $student$& 33.41\\
    Our $student$ ($D$ with features)& 33.40\\
    \quad\quad\quad+ dropout on $student$& 32.45\\
    \end{tabular}}
    {\caption{{\small {\bf CIFAR-100 Evaluation} The component evaluation is presented. We use the same $teacher$ and $student$ models as in CIFAR-10. ResNet-164 Parameters: \textbf{2.6M}, FLOPs: \textbf{97.49B}. ResNet-20 Parameters: \textbf{0.27M}, FLOPs: \textbf{10.52B}.}}\label{tab:CIFAR100}}
  \end{floatrow}
\end{table*}%

\textbf{CIFAR-10, Table~\ref{tab:CIFAR10}.} All compression baselines, based on $student$ with ResNet-20, have only around $1.5\%$ drop in performance compared to the $teacher$ (ResNet-164). Moreover, they are all better than the $student$ network, trained with supervision (i.e.~labels), which is $2\%$ behind the $teacher$. Our complete model benefits from the dropout on the adversarial samples and achieves the best performance using feature input to $D$.

\begin{table*}[h!]
  \floatsetup{floatrowsep=qquad, captionskip=4pt}
  \begin{floatrow}[2]
    \ttabbox%
    {\begin{tabular}{l|c}
    \hline
    Model & Error[\%]\\\hline
    Supervised $teacher$& 3.98\\
    ResNet-164\\\hline
    Supervised $student$& 4.20\\
    ResNet-20\\\hline
    Our $student$ ($D$ with logits)& 3.74\\
    \quad\quad\quad+ dropout on $student$& 3.81\\
    Our $student$ ($D$ with features)& 3.74\\
    \quad\quad\quad+ dropout on $student$& 3.66\\
    \end{tabular}}
     {\caption{{\small {\bf SVHN Evaluation} We evaluate the components of our approach. ResNet-164 Parameters: \textbf{2.6M}, FLOPs: \textbf{97.49B}. ResNet-20 Parameters: \textbf{0.27M}, FLOPs: \textbf{10.52B}. The $teacher$ and $student$ model are similar to CIFAR evaluation.}}\label{tab:SVHN}}
    \hfill%
    \ttabbox%
   {\begin{tabular}{l|c}
    \hline
    Model & Error[\%]\\\hline
    Supervised $teacher$& 7.98\\
    NiN&\\\hline
    Supervised $student$& 8.77\\
    LeNet-4&\\\hline
    Our $student$ ($D$ with logits)& 8.90\\
    \quad\quad\quad+ dropout on $student$& 8.84\\
    Our $student$ ($D$ with features)& 8.86\\
    \quad\quad\quad+ dropout on $student$& 8.61\\
\end{tabular}}
     {\caption{{\small {\bf Fashion-MNIST Evaluation} We evaluate the components with different $teacher$ and $student$. NiN Parameters: \textbf{10.6M}, FLOPs: \textbf{60.23B}. LeNet-4 Parameters: \textbf{2.3M}, FLOPs: \textbf{7.06B}. Our $student$, LeNet-4, has around 5x less parameters.}}\label{tab:FMNIST}}
  \end{floatrow}
\end{table*}%

\begin{table*}[h!]
\centering
\caption{{\footnotesize {\bf CIFAR-10 and CIFAR-100 Comparisons} We compare our results and number of network parameters with related methods on similar architectures. We use ResNet-20 for our $student$ and our complete model.}}
\begin{tabular}{c|c|c||c|c|c}
\textbf{CIFAR-10} & Error[\%]& Param. & \textbf{CIFAR-100} & Error[\%]& Param.\\\hline
L2 - Ba {\it et al.}~\cite{ba2014deep}& 9.07 & 0.27M  & Yim {\it et al.}~\cite{yim2017gift} & 36.67&-\\
Hinton {\it et al.}~\cite{hinton2015distilling}& 8.88 & 0.27M & FitNets~\cite{fitnets} & 35.04 & 2.50M\\
Quantization~\cite{zhu2016trained} & 8.87&0.27M & Hinton {\it et al.}~\cite{hinton2015distilling}& 33.34 & 0.27M \\
FitNets~\cite{fitnets}& 8.39&2.50M&L2 - Ba {\it et al.}~\cite{ba2014deep}& 32.79&0.27M\\
Binary Connect~\cite{courbariaux2015binary}& 8.27&15.20M&Our $student$& \textbf{32.45} & 0.27M\\
Yim {\it et al.}~\cite{yim2017gift} & 11.30&-& & &\\
Our $student$& \textbf{8.08}&0.27M& & &\\
\end{tabular}
\label{tab:CIFARcomp}
\end{table*}

\textbf{CIFAR-100, Table~\ref{tab:CIFAR100}.} We also use Resnet-164 for $teacher$ and ResNet-20 for $student$ to have 10x less parameters as in CIFAR-10. In this evaluation, the performance drop between the $teacher$ and the compressed models is slightly larger. The overall behavior is similar to CIFAR-10. However, the error is reduced by $1\%$ after adding the dropout to the $student$ using features as input to $D$. Here, we had the biggest improvement after using dropout.

\textbf{SVHN, Table~\ref{tab:SVHN}.} In this experiment, the $teacher$ and $student$ architectures are still the same. Although, we tried the Network in Network (NiN) and LeNet-4 as teacher and $student$, the pair did not perform as well as ResNet. Unlike in the previous experiments, here the Adam optimizer was used, as it improved across all ablation results. Notice that our $student$ achieves better performance not only from the same network trained with labels, the supervised $student$, but from the teacher too. This is a known positive side product of the distillation~\cite{ba2014deep}.

\textbf{Fashion-MNIST, Table~\ref{tab:FMNIST}.} We select Network in Network (NiN) as $teacher$ and LeNet-4 as $student$. The dataset is relative simple and thus a ResNet architecture is not necessary. All approaches are close to each other.It is clear that the features input to $D$ and the dropout are important to obtain the best performance in comparison to the other baselines. For instance, the $student$ network trained with supervision (error $8.77\%$) is better than our baselines other than our complete model (error $8.61\%$). Finally the Adam optimizer has been used.
\begin{table*}[t]
\centering
\caption{{\footnotesize {\bf SVHN and F-MNIST Comparisons} We compare our results and number of network parameters with related methods on the same $student$ architecture that is ResNet-20 for SVHN and LeNet-4 for Fashion-MNIST.}}
\begin{tabular}{c|c|c||c|c|c}
\textbf{SVHN} & Error[\%]& Param. & \textbf{F-MNIST} & Error[\%]& Param.\\\hline
L2 - Ba {\it et al.}~\cite{ba2014deep}& 3.75 & 0.27M  & L2 - Ba {\it et al.}~\cite{ba2014deep} & 8.89& 2.3M\\
Hinton {\it et al.}~\cite{hinton2015distilling}& 3.66 & 0.27M &Hinton {\it et al.}~\cite{hinton2015distilling}& 8.71 & 2.3M\\
Our $student$& \textbf{3.66}&0.27M & Our $student$& \textbf{8.64} &  2.3M \\
\end{tabular}
\label{tab:SVHNFMNISTcomp}
\end{table*}

\textbf{Common conclusions} There is a number of common outcomes for all evaluations: 1.~the adversarial compression reaches the lowest error when using features as input to $D$; 2.~our $student$ performs always better than training the same network with labels (i.e.~supervised student) and 3.~we achieve good generalization on different $teacher$ - $student$ architectures.

\textbf{Comparisons to state-of-the-art} In Tables~\ref{tab:CIFARcomp} and~\ref{tab:SVHNFMNISTcomp}, we compare our $student$ with other compression strategies on CIFAR-10 and CIFAR-100. We choose four distillation~\cite{ba2014deep,hinton2015distilling,fitnets,yim2017gift} and two quantization~\cite{courbariaux2015binary,zhu2016trained} approaches for CIFAR-10. We examine the same four distillation methods for a comparison on CIFAR-100. The work of Ba and Caruana~\cite{ba2014deep} is closer to our approach, because it relies on L2 minimization, though it is on the logits (see  Table~\ref{tab:CIFARcomp}). The Knowledge Distillation (KD)~\cite{hinton2015distilling} is also related to our idea, but it relies on labels. Both evaluations demonstrate that we achieve the lowest error and our $student$ has the smallest number of parameters. In addition, we compare our results on SVHN and Fashion-MNIST with two distillation approaches (see Table~\ref{tab:SVHNFMNISTcomp}). The error here is much lower for methods, but we are consistently better than the other approaches. Next, We demonstrate the same findings on large-scale experiments.

\subsection{ImageNet Evaluation} \label{sec:imagenet}
We perform an evaluation on ImageNet to examine whether the distillation is possible on a large-scale dataset with class number set to 1000. The $teacher$ is a pre-trained ResNet-152, while we try two different $student$ architectures. At first, we choose ResNet-18 to train our $student$ using features as input to $D$ and adding the dropout on the adversarial samples. Regarding the experimental settings, we have set the batch size to 256, while the rest hyper-parameters remain the same. All networks use the the average pool layer to output features for $D$. We evaluate on the validation dataset. The results are presented in Table~\ref{tab:ImagNet}. 

Our findings are consistent with the earlier evaluations. Our best performing model (features input to $D$ and dropout on $student$) perform at best and better than the student trained with supervision. Secondly, we examine a stronger $student$ where we employ ResNet-50 for training our model. We present our results in Table~\ref{tab:ImagNet2} where we also compare with binarization, distillation and factorization methods. Although we achieve the best results, MobileNets has fewer parameters. We see the adversarial network compression on MobileNets as future work.

\begin{table*}[h!]
\centering
\caption{{\footnotesize {\bf ImageNet Baselines} We evaluate the components of our approach. ResNet-152 Parameters: \textbf{58.21M}, FLOPs: \textbf{5587B}. ResNet-18 Parameters: \textbf{13.95M}, FLOPs: \textbf{883.73B}. Our $student$ has around 4x less parameters. Our model has features as input to $D$ and dropout on the adversarial samples.}}
\begin{tabular}{l|cc}
\hline
 & Top-1& Top-5\\
Model&Error[\%]&Error[\%] \\\hline
Supervised $teacher$ (ResNet-152) & 27.63&5.90\\
Supervised $student$ (ResNet-18) & 43.33 & 20.11\\\hline
Our $student$ ($D$ with features)& 33.31 & 11.96 \\
\quad\quad\quad+ dropout on $student$& \textbf{32.89} & \textbf{11.72}\\
\end{tabular}
\label{tab:ImagNet}
\end{table*}

\begin{table*}[h!]
\centering
\caption{{\footnotesize {\bf ImageNet Evaluation} We evaluate two versions of our $student$ and compare with related methods. ResNet-152 Parameters: \textbf{58.21M}, FLOPs: \textbf{5587B}. ResNet-50 Parameters: \textbf{37.49M}, FLOPs: \textbf{2667B}. Our $student$, ResNet-50, has around 2x less parameters. We also include the $student$ ResNet-18 from the evaluation in Table~\ref{tab:ImagNet}. Our $student$ is trained with features as input to $D$ and dropout on the $student$.}}
\begin{tabular}{l|ccc}
\hline
Model& Top-1 Error[\%]& Top-5 Error[\%]& Parameters\\
Supervised $teacher$ (ResNet-152) & 27.63&5.90&58.21M\\
Supervised $student$ (ResNet-50) & 30.30& 10.61&37.49M\\\hline
XNOR~\cite{rastegari2016xnor} (ResNet-18)& 48.80 &26.80 & 13.95M\\
Binary-Weight~\cite{rastegari2016xnor} (ResNet-18) & 39.20 & 17.00&13.95M\\
L2 - Ba {\it et al.}~\cite{ba2014deep} (ResNet-18) & 33.28 & 11.86&13.95M \\
MobileNets~\cite{howard2017mobilenets} & 29.27 &10.51&4.20M \\
L2 - Ba {\it et al.}~\cite{ba2014deep} (ResNet-50) & 27.99&9.46&37.49M \\
Our $student$ (ResNet-18)& 32.89 & 11.72 &13.95M \\
Our $student$ (ResNet-50)& \textbf{27.48} & \textbf{8.75}&37.49M \\ 
\end{tabular}
\label{tab:ImagNet2}
\end{table*}

\section{Conclusion}
We have presented the adversarial network compression for knowledge transfer between a large model and a smaller one with limited capacity. We have empirically shown that regularization helps the student to compete with the discriminator. Finally, we show state-of-the-art performance without using labels in an extensive evaluation of five datasets, three teacher and three student architectures. As future work, we aim to explore adversarial schemes with more discriminators that use intermediate feature representations, as well as, transferring our approach to different tasks such as object detection and segmentation.

\section{Acknowledgements}
This research was partially supported by BMWi - Federal Ministry for Economic Affairs and Energy (MEC-View Project).

\bibliographystyle{splncs04}
\bibliography{egbib}
\end{document}